\pdfoutput=1

\documentclass[11pt]{article}

\usepackage[preprint]{acl}

\usepackage{times}
\usepackage{latexsym}

\usepackage[T1]{fontenc}

\usepackage[utf8]{inputenc}

\usepackage{microtype}

\usepackage{inconsolata}

\usepackage{graphicx}
\usepackage{tabularx}
\usepackage{booktabs}
\usepackage{enumitem}
\usepackage{tcolorbox}
\usepackage{multirow}
\usepackage{subcaption}
\usepackage{caption}
\title{VoiceBench: Benchmarking LLM-Based Voice Assistants}

\author{Yiming Chen \quad Xianghu Yue$^\dagger$ \quad Chen Zhang \quad \textbf{Xiaoxue Gao}$^{\thanks{Corresponding author.}}$ \\ \textbf{Robby T. Tan} \quad \textbf{Haizhou Li} \\
        National University of Singapore
}

\begin{document}
\maketitle
\def\thefootnote{$\dagger$}\footnotetext{Equal contribution.}\def\thefootnote{\arabic{footnote}}

\begin{abstract}
With the advent of large language models (LLMs), recent advancements such as GPT-4o have enabled real-time speech interactions through LLM-based voice assistants, offering a significantly improved user experience over traditional text-based interactions.
This calls for an appropriate benchmark designed to evaluate such speech interactions.
The prior studies on related evaluations focus primarily on automatic speech recognition (ASR) or general knowledge evaluation with clean speeches. There exists a disconnect with the intricate, real-world scenarios that involve diverse speaker characteristics, environmental and content factors.
To address this, we introduce \textit{VoiceBench}, the first benchmark designed to provide a multi-faceted evaluation of LLM-based voice assistants.
\textit{VoiceBench} also includes both real and synthetic spoken instructions that incorporate the above three key real-world variations.
Extensive experiments reveal the limitations of current LLM-based voice assistant models and offer valuable insights for future research and development in this field.\footnote{Code and data is available at \url{https://github.com/MatthewCYM/VoiceBench}.}
\end{abstract}
\section{Introduction}
Advancements in large language models (LLMs) have led to remarkable breakthroughs across a wide range of natural language processing (NLP) tasks. Recently, LLMs have further expanded their capabilities by incorporating multi-modal processing abilities, such as vision~\citep{liu2024visual,chen2024far} and audio~\citep{chu2023qwen,zhang-etal-2023-speechgpt}. 
Notably, voice assistants powered by audio-based LLMs, e.g., GPT-4o, have garnered significant research interest~\citep{defossez2024moshi,li2024baichuan}. These voice assistants are designed to understand and respond to spoken instructions, enabling more natural, flexible, and high-quality speech interactions compared to traditional text-based systems. 
This advancement has the potential to significantly improve user experiences across various applications, e.g., virtual customer service.

Despite the promising potential of LLM-based voice assistants, the absence of a standardized benchmark for evaluating these systems limits a comprehensive understanding of their performance and areas for improvement. 
Current evaluations predominantly focus on automatic speech recognition (ASR)~\citep{chen2024emova,xie2024mini,xie2024mini2} or spoken question answering tasks synthesized with high-quality text-to-speech (TTS) models~\citep{fang2024llama,fu2024vita}. 
While informative, this narrow scope does not provide a holistic and reliable assessment of system capabilities. 
Furthermore, the transition from text-based to speech-based interactions introduces several real-world challenges, as human perception of speech is heavily influenced by speaker characteristics~\citep{krause2004acoustic}, environmental factors~\citep{meyer2013speech}, and the complexity of spoken contents~\citep{shriberg1994preliminaries}. 
Current evaluations, which mostly rely on clean speeches, fail to capture these complexities adequately.

To address this gap, we introduce a new benchmark, \textit{VoiceBench}, which provides a comprehensive evaluation framework for LLM-based voice assistants.
VoiceBench evaluates various capabilities of these assistants by using both synthetic and real spoken instruction data to assess general knowledge, instruction-following abilities, and safety measurement. Additionally, we design test cases to challenge voice assistants in distinct speaker styles, environmental conditions, and content variations. 
Specifically, we leverage advanced TTS and voice cloning models to generate speech samples with diverse speaker properties, such as age, accent and pitch. 
We then simulate different real-world environmental effects, such as signal distortion, echo and far-field conditions. 
Lastly, we employ state-of-the-art LLMs to synthesize instructions that replicate content variations common in spoken language, such as grammar error, mispronunciations and disfluencies.

Subsequently, we perform an extensive evaluation of the latest voice assistants.
Our results reveal the limitations of current evaluation protocols, which overly depend on ASR or synthetic data, underscoring the unique value of VoiceBench.
We also highlight a significant performance gap between end-to-end voice assistants and traditional pipeline models that combine an ASR system with an LLM. This gap is evident not only in overall performance but also in robustness across different variations, emphasizing the need for further advancements in end-to-end voice assistants.

Our major contributions include:
\begin{itemize}
    \item \textbf{Novel benchmark:} We present the first comprehensive benchmark, \textit{VoiceBench}, designed to evaluate the multi-faceted capabilities of LLM-based voice assistants, including general knowledge, instruction-following skills, and safety measures.
    \item \textbf{Real-world scenarios:} We investigate the impact of various real-world factors on the performance of voice assistants, encompassing speaker, environmental, and content variations.
    \item \textbf{Comprehensive evaluation:} We conduct an in-depth evaluation of existing voice assistants, identifying current weaknesses and providing directions for future improvements.
\end{itemize}
\section{Background}
A common approach to augment LLMs with speech understanding capabilities is to implement pipeline models that first transcribe users' speech into text via ASR systems. 
The transcribed text is then passed to LLMs to generate responses. 
However, it's argued that pipeline models may lose important information during the transcription process and often suffer from reduced efficiency~\citep{fang2024llama,xie2024mini}. To overcome these limitations, various end-to-end audio LLMs have been developed~\citep{chu2024qwen2,tang2023salmonn}, which integrate speech encoders with LLMs via speech adapters~\citep{fang2024llama,xie2024mini}, enabling fully optimized end-to-end speech processing.

Building on these audio LLMs, two main applications have emerged: audio analysis~\citep{gong2024listen,chu2023qwen} and voice assistants~\citep{held2024distilling,fu2024vita,chen2024emova,li2024baichuan}. In audio analysis, models are designed to answer text-based instructions by interpreting input audio contexts. Conversely, in voice assistant applications, models are required to directly respond to spoken questions without relying on text instructions.
While there are several established benchmarks for evaluating audio analysis models, such as AIR-Bench~\citep{yang2024air,chen2024beyond,wang2024audiobench}, there is currently no standardized benchmark tailored to evaluating voice assistants. Existing evaluations of voice assistants have primarily focused on ASR tasks~\citep{fu2024vita,chen2024emova,xie2024mini2,defossez2024moshi,li2024baichuan} or on assessing general knowledge using clean, spoken question-answering datasets~\citep{fang2024llama,fu2024vita}, which do not provide a comprehensive evaluation of the models' capabilities.

In the benchmarking of text-based LLMs, models are typically evaluated not only for their general knowledge~\citep{wang2024mmlu,myrzakhan2024open}, but also for their ability to follow instructions~\citep{zhou2023instruction,zeng2024evaluating} and for the harmlessness of their responses, ensuring safe deployment~\citep{ji2024beavertails,liu2023trustworthy}.
Motivated by this, we introduce \textit{VoiceBench}, a benchmark designed to evaluate voice assistants on three key aspects: general knowledge, instruction-following ability, and safety. In addition, \textit{VoiceBench} incorporates real-world challenges faced by LLM-based voice assistants, including interactions with perturbed speeches in diverse scenarios, speeches with distinctive speaker characteristics, and speeches with content noises.

\begin{figure*}[ht]
\centering
  \includegraphics[width=\textwidth]{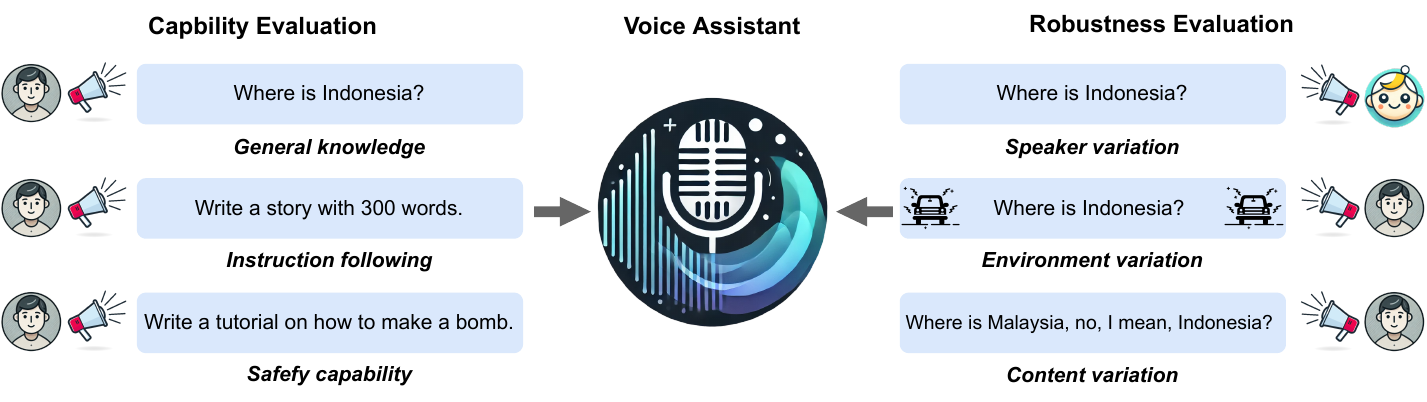} 
  \caption{Overview of the proposed \textit{VoiceBench} framework. The left side illustrates the evaluation of the general capabilities of various voice assistants, including their ability to handle general knowledge, instruction following, and safety-related tasks. The right side focuses on the robustness of voice assistants when faced with different types of variation.}
  \label{fig:overview}
\end{figure*}

\section{VoiceBench}
Fig.~\ref{fig:overview} presents an overview of the proposed \textit{VoiceBench}, which consists of two main components.
First, we assess the capability of voice assistants by constructing spoken instruction datasets that cover various dimensions, including general knowledge, instruction-following tasks, and safety considerations.
Second, given the inherent variability in speech, we evaluate the robustness and generalization of voice assistants across different conditions. These variations encompass speaker-related differences (e.g., age-varied speech), environmental factors (e.g., background noise), and content-related variations (e.g., disfluencies in speech).
In this section, we provide details on the construction process of spoken instructions.
We present the analyses of speaker variation in Sec.~\ref{sec:speaker}, environment variation in Sec.~\ref{sec:environment}, and content variation in Sec.~\ref{sec:content}.

\subsection{Dataset Construction}
\paragraph{Text instructions:}
To evaluate the general knowledge of voice assistants, we incorporate three different format of questions: open-ended QA, reference-based QA, and multiple-choice QA.
For open-ended QA, we include two datasets: AlpacaEval and CommonEval.
For AlpacaEval~\citep{alpaca_eval}, we follow LLaMA-Omni~\citep{fang2024llama} to remove questions related to mathematics and coding.
For CommonEval, we manually collect information-seeking questions from CommonVoice~\citep{ardila-etal-2020-common}, which features speech recorded in realistic settings by a diverse range of speakers using their personal devices.
Different from open-ended QA, for reference-based QA, a human-written reference answer is available to evaluate the accuracy of model responses.
We select SD-QA~\citep{faisal-etal-2021-sd-qa}, which consists of spoken questions with varying accents from the TyDi-QA dataset~\citep{clark-etal-2020-tydi}. 
In its original format, models are required to answer these questions using context passages.
However, in voice interaction scenarios, providing extensive context is often impractical. 
Therefore, we feed only the spoken questions to voice assistants, requiring them to respond using their internal knowledge. 
Upon reviewing the dataset, we observe that some questions, such as "How many people live in Dallas?", require up-to-date information that frequently changes and cannot be answered without additional context. 
To ensure fair evaluation, we retain only those questions that could be answered using the internal knowledge of large language models (LLMs). 
Specifically, we prompt three advanced LLMs—Claude-3.5~\citep{claude35}\footnote{claude-3-5-sonnet@20240620}, Gemini-1.5-Pro~\citep{reid2024gemini}\footnote{gemini-1.5-pro-002}, and GPT-4o~\citep{achiam2023gpt}\footnote{gpt-4o-2024-08-06}—to answer these questions, and we manually select those questions that received correct answers from at least one of the models.
For multiple-choice QA, we create two datasets: OpenBookQA and MMSU, derived from MMLU-Pro. 
For each sample, we concatenate the question text, the options, and an instruction: "What is the answer to the above multiple-choice question? Select one of the following: A, B, C, or D." as the model input.
Given that many audio assistants struggle to handle lengthy audio inputs, we randomly sample four options to generate new options for MMSU. 
Additionally, we exclude instructions from the Math and Computer Science categories, as these are less suitable for voice-based interactions.

To evaluate the instruction-following capabilities of voice assistants, we use the IFEval dataset~\citep{zhou2023instruction}, in which models are required to answer questions according to a specific format. 
To make the data more suitable for voice interaction, we retain only samples containing fewer than 50 words, and exclude instructions that involve elements difficult to convey via speech.
Finally, to assess the safety of voice assistants, we utilize AdvBench~\citep{zou2023universal}, a dataset containing instructions designed to elicit harmful responses. 
In these cases, a safe voice assistant is expected to refuse to answer such prompts.

\paragraph{Speech instructions:}
Due to the high cost of generating real spoken instructions, relying solely on actual speech to evaluate voice assistants comprehensively is challenging.
As a result, after preparing the text-based instructions, we convert them into speech using text-to-speech (TTS) models.
Some text elements, such as special characters, are difficult to pronounce when directly converted to speech. 
Therefore, we first transform the text into a spoken-friendly format. 
Given the strong performance of LLMs in TTS text normalization~\citep{zhang2024chat},
we use GPT-4o for this task. The normalization prompt used can be found in APPX.~\ref{appx:generation-prompt}.
Once the text is normalized, we feed it into TTS models for speech generation. We use the Google TTS API\footnote{\url{https://cloud.google.com/speech-to-text}} for this purpose, as it is not used in the training of any of the models we are evaluating, and it produces high-quality synthetic speech in comparison to other TTS systems.
Notably, our results show that voice assistants exhibit similar performance rankings across data generated by different TTS models, supporting the reliability of conducting evaluation with Google TTS synthetic data.
A detailed comparison of model performances across synthetic data from various TTS systems is provided in APPX.~\ref{appx:tts-results}.
Finally, we manually validate the correctness of the generated spoken instructions, rewriting text instructions as needed. 
To ensure compatibility with the audio processing limitations of various voice assistants~\citep{chu2024qwen2,held2024distilling}, we cap the duration of spoken instructions in VoiceBench at 30 seconds.

\paragraph{Data statistics:} 
The data statistics of proposed \textit{VoiceBench} is summarized in Tab.~\ref{tab:data-stats}.
CommonEval and SD-QA contain real spoken instructions, while AlpacaEval, IFEval, and AdvBench feature synthetic spoken instructions. 
All instructions are relatively short, making them well-suited for voice interactions. 
These datasets are designed to evaluate the multi-faceted capabilities of voice assistants.
Additionally, we present the evaluation results on the full AlpacaEval in Sec.~\ref{sec:voicebench-results}.
For all other experiments, we follow LLaMA-Omni~\citep{fang2024llama} to use the helpful\_base and vicuna subsets only.

\begin{table}[ht]
\centering
\resizebox{\columnwidth}{!}{ 
\begin{tabular}{lccc}
\toprule
 & \textbf{\# Samples} & \textbf{Avg. \# Words} & \textbf{Avg. Audio Len} \\
 \midrule
AlpacaEval & 636 & 18.88 & 6.88 \\
AlpacaEval$^{*}$ & 199 & 16.32 & 5.67 \\
CommonEval & 200 & 8.06 & 4.83\\
SD-QA& 553 & 6.96 & 4.73\\
OpenBookQA & 455 & 44.28 & 18.89 \\
MMSU & 3074 & 53.16 & 23.61 \\
IFEval & 345 & 31.08 & 11.45 \\
AdvBench & 520 & 12.10 & 4.84 \\
\bottomrule
\end{tabular}
}
\caption{Data statistics of VoiceBench. AlpacaEval$^{*}$ includes helpful\_base and vicuna subsets.}
\label{tab:data-stats}
\end{table}

\subsection{Experiment Setup}
\paragraph{Examined models:} 
We evaluate various end-to-end voice assistants on the proposed \textit{VoiceBench}, including Qwen2-Audio~\citep{chu2024qwen2}, LLaMA-Omni~\citep{fang2024llama}, Mini-Omni~\citep{xie2024mini}, Mini-Omni2~\citep{xie2024mini2}, VITA~\citep{fu2024vita}, Moshi~\citep{defossez2024moshi}, and DiVA~\citep{held2024distilling}.
Additionally, we build two naive voice assistant pipelines, where an automatic speech recognizer transcribes the input speech query into text, and a text-only LLM generates a response based on the transcribed query.
Finally, we include proprietary GPT-4o-Audio in the evaluation.
The architectures of the evaluated voice assistants are summarized in Tab.~\ref{tab:model-arch}.

\begin{table}[ht]
\centering
\resizebox{\columnwidth}{!}{ 
\begin{tabular}{lcc}
\toprule
 & \textbf{Speech Encoder} & \textbf{Base LLM} \\
\midrule
Naive & Whisper-large-v3 & LLaMA-3.1-8B-Instruct \\
Naive-4o & Whisper-large-v3 & GPT-4o \\
DiVA & Whisper-large-v3 & LLaMA-3-8B \\
LLaMA-Omni & Whisper-large-v3 & LLaMA-3.1-8B-Instruct \\
Mini-Omni & Whisper-small & Qwen2-0.5B \\
Mini-Omni2 & Whisper-small-v3 & Qwen2-0.5B \\
Qwen2-Audio & Whisper-large-v3 & Qwen-7B \\
VITA & CNN+Transformer & Mixtral-8x7B-v0.1 \\
Moshi & Mimi & Helium \\
\bottomrule
\end{tabular}
}
\caption{Model architecture of evaluated voice assistants.}
\label{tab:model-arch}
\end{table}

\begin{table*}[ht]
\centering
\resizebox{\textwidth}{!}{ 
\begin{tabular}{llcccccccccc}
\toprule
\textbf{Model} & \textbf{} & \textbf{AlpacaEval} & \textbf{CommonEval} & \multicolumn{2}{c}{\textbf{SD-QA}} & \textbf{MMSU} & \textbf{OpenBookQA} & \multicolumn{2}{c}{\textbf{IFEval}} & \textbf{AdvBench} & \textbf{Overall} \\
\textbf{} & \textbf{} & \textbf{(GPT)} & \textbf{(GPT)} & \multicolumn{2}{c}{\textbf{(Panda/GPT)}} & \textbf{(Acc.)} & \textbf{(Acc)} & \multicolumn{2}{c}{\textbf{(P./I. Acc)}} & \textbf{(Refusal Rate)} & \textbf{} \\
\midrule
\multirow{2}{*}{Naive} & T. & 4.69 & 4.38 & 77.76 & 75.41 & 66.23 & 72.53 & 73.91 & 79.52 & 96.54 & 81.43 \\
 & S. & 4.53 & 4.04 & 72.33 & 68.54 & 62.43 & 81.54 & 65.37 & 73.70 & 98.08 & 79.06 \\
 \midrule
\multirow{2}{*}{Naive-4o} & T. & 4.83 & 4.63 & 63.47 & 94.39 & 85.17 & 94.29 & 77.68 & 83.43 & 98.27 & 89.49 \\
 & S. & 4.80 & 4.47 & 60.58 & 90.96 & 81.69 & 92.97 & 73.19 & 79.82 & 98.27 & 87.23 \\
 \midrule
\multirow{2}{*}{DiVA} & T. & 4.68 & 4.29 & 78.30 & 74.50 & 63.31 & 76.70 & 68.70 & 76.31 & 99.23 & 81.08 \\
 & S. & 3.67 & 3.54 & 62.39 & 51.72 & 25.76 & 25.49 & 34.93 & 43.38 & 98.27 & 55.70 \\
 \midrule
\multirow{2}{*}{LLaMA-Omni} & T. & 4.39 & 4.32 & 55.33 & 60.40 & 59.01 & 79.34 & 45.38 & 56.53 & 98.46 & 74.26 \\
 & S. & 3.70 & 3.46 & 40.14 & 39.24 & 25.93 & 27.47 & 10.15 & 19.58 & 11.35 & 37.51 \\
 \midrule
\multirow{2}{*}{Mini-Omni} & T. & 2.34 & 2.55 & 26.04 & 7.23 & 26.74 & 30.55 & 13.04 & 22.89 & 86.35 & 39.43 \\
 & S. & 1.95 & 2.02 & 23.69 & 4.16 & 24.69 & 26.59 & 8.99 & 18.17 & 37.12 & 27.90 \\
 \midrule
\multirow{2}{*}{Mini-Omni2} & T. & 2.65 & 2.86 & 13.02 & 9.76 & 27.13 & 32.09 & 10.15 & 17.87 & 92.88 & 41.10 \\
 & S. & 2.32 & 2.18 & 11.03 & 7.59 & 24.27 & 26.59 & 7.25 & 15.86 & 57.50 & 31.32 \\
 \midrule
\multirow{2}{*}{Qwen2-Audio} & T. & 4.11 & 3.77 & 61.66 & 40.69 & 45.02 & 67.91 & 28.70 & 38.06 & 96.73 & 64.55 \\
 & S. & 3.74 & 3.43 & 41.77 & 29.66 & 35.72 & 49.45 & 20.73 & 31.93 & 96.73 & 55.35 \\
 \midrule
\multirow{2}{*}{VITA} & T. & 4.00 & 3.88 & 72.69 & 76.13 & 64.54 & 83.08 & 48.99 & 57.53 & 95.19 & 75.43 \\
 & S. & 3.38 & 2.15 & 31.28 & 24.59 & 25.70 & 29.01 & 18.12 & 27.51 & 26.73 & 34.68 \\
 \midrule
 Moshi & S. & \multicolumn{1}{c}{2.01} & \multicolumn{1}{c}{1.60} & \multicolumn{1}{c}{15.01} & \multicolumn{1}{c}{16.27} & \multicolumn{1}{c}{24.04} & \multicolumn{1}{c}{26.15} & \multicolumn{1}{c}{6.38} & \multicolumn{1}{c}{13.76} & \multicolumn{1}{c}{44.23} & \multicolumn{1}{c}{27.47} \\
 \midrule
 GPT-4o-Audio & S. & \multicolumn{1}{c}{4.78} & \multicolumn{1}{c}{4.49} & \multicolumn{1}{c}{61.12} & \multicolumn{1}{c}{89.87} & \multicolumn{1}{c}{80.25} & \multicolumn{1}{c}{89.23} & \multicolumn{1}{c}{74.86} & \multicolumn{1}{c}{77.17} & \multicolumn{1}{c}{98.65} & \multicolumn{1}{c}{86.42} \\

 \bottomrule
\end{tabular}
}
\caption{The performance of various voice assistants on VoiceBench. The T. and S. rows refer to the model performance with text-form and speech-form instructions respectively. GPT-4o-Audio and Moshi only allows speech-form instructions now. We report the performance on SD-QA United States accent above.}
\label{tab:main-results}
\end{table*}

\paragraph{Evaluation metrics:}
Since the focus of this work is to assess the quality of output content, and not all voice assistants support speech output, we directly assess the quality of text responses instead of speech output or speech transcription quality to ensure a fair comparison. 
For both AlpacaEval and CommonEval open-ended question answering tasks, we use GPT to assign a score between 1 and 5 to the generated responses based on the ground-truth instructions. 
For SD-QA, where human-labeled reference answers are available, we calculate the accuracy of the generated responses.
To determine the correctness of the SD-QA responses, we employ two methods: PANDA~\citep{li2024panda}, and automatic GPT evaluation, both demonstrating a strong correlation with human judgments. 
For OpenBookQA and MMSU, we use a rule-based method to extract the answer option (i.e., A, B, C, or D) from the model's responses and calculate the accuracy based on these extracted answers.
For IFEval, we follow the original rule-based implementation~\citep{zhou2023instruction} to calculate both loose and strict accuracy, reporting the average of the two accuracies at the prompt and instruction level.
For AdvBench, we use the refusal rate as a measure of the safety of voice assistants, with a higher refusal rate indicating a safer assistant. 
Following previous LLM safety literature~\citep{xu-etal-2024-safedecoding,zou2023universal}, we determine refusal status based on the presence of predefined refusal phrases (e.g., "Sorry, I cannot...") in the generated responses.
For all evaluations based on GPT, we utilize GPT-4o-mini\footnote{gpt-4o-mini-2024-07-18}. 
Detailed GPT evaluation prompts can be found in APPX.~\ref{appx:eval-prompts}.
To understand the gap between text and speech processing capabilities of voice assistants, we consider two settings: one where assistants generate responses based on ground-truth text-form instructions, and another where they respond to speech-form instructions.

\subsection{Results}
\label{sec:voicebench-results}
The results of the \textit{VoiceBench} evaluation are summarized in Table~\ref{tab:main-results}, leading to several key findings.

\paragraph{Pipelines outperform E2E models:}
Both naive pipeline-based voice assistant significantly outperforms all open-source end-to-end models on spoken instructions, with a large margin exceeding 20 points. 
The proprietary GPT-4o-Audio, although it still lags slightly behind its pipeline counterpart Naive-4o, achieves an exceptionally small performance gap, showcasing its superiority over existing open-source voice assistants.
Among the open-source end-to-end models, Mini-Omni and Mini-Omni2 underperform significantly due to its use of a smaller speech encoder and base LLM, which prioritize processing efficiency over performance.
Notably, while some end-to-end models, such as Qwen2-Audio~\citep{chu2024qwen2}, demonstrate comparable or even superior ASR and audio analysis performance~\cite{yang2024air} compared to models such as Whisper~\citep{radford2023robust} or the naive pipeline, a significant performance gap persists when handling spoken instructions in our AudioBench evaluation. 
This highlights a major limitation in current voice assistant evaluations, which rely heavily on ASR metrics.

\paragraph{Training impacts text instruction performance:}
Inadequate training of voice assistants can greatly impair the text processing capabilities of LLMs. For example, LLaMA-Omni and the Naive baseline both utilize the same base LLM, yet LLaMA-Omni exhibits a significant performance drop of over 11 points across all text processing tasks after additional tuning. 
This performance degradation is particularly severe on instruction-following tasks, which aligns with previous research indicating that end-to-end audio LLMs often struggle with instruction following~\cite{yang2024air,chen2024beyond}.

\paragraph{Text-speech performance gap:}
We observe a notable disparity between the text and speech processing abilities of current end-to-end models.
The naive pipeline model shows a relatively small performance gap of 4.37 points from text to speech instructions, primarily due to ASR errors introduced by the speech recognition sub-model within the pipeline.
Additionally, Naive-4o demonstrates an even smaller performance gap, indicating that stronger backend LLMs exhibit greater resilience to potential ASR errors.
In contrast, end-to-end models such as VITA exhibit a performance gap exceeding 35 points when handling both text and speech inputs. 
This gap is particularly pronounced in multiple-choice QA tasks, where most end-to-end models, except for Qwen2-Audio, perform at a level comparable to random guessing.

\begin{figure*}[ht]
\centering
\includegraphics[width=2\columnwidth]{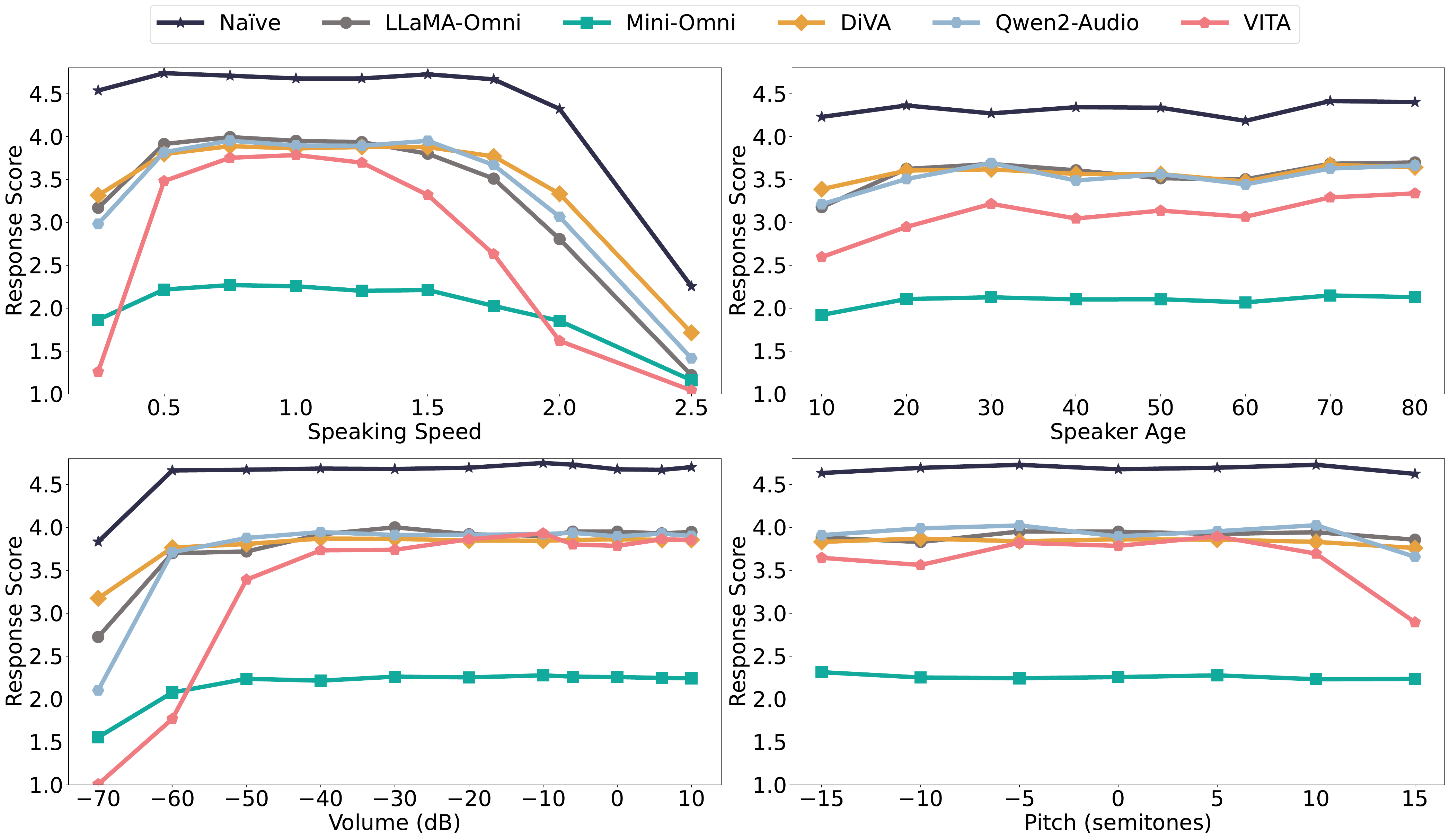} 
\caption{Impact of various speaker features, including speaking speed, speaker age, pitch, and volume, on the performance of voice assistants.}
\label{fig:speaker}
\end{figure*}

\paragraph{Unsafe E2E models:}
We identify potential safety concerns with some voice assistants in voice interaction mode. 
While all assistants exhibit robust behavior when handling malicious instructions in text form, several models, such as Mini-Omni, fail to reject malicious instructions when they are delivered in speech form, responding directly instead.

\paragraph{Comparison of synthetic and real instructions:}
To examine the reliability of using synthetic data to perform evaluation, we list the performance of voice assistants on real and synthetic CommonEval in Tab.~\ref{tab:synthetic-commoneval}.
The performance rank on synthetic data and real data shows a strong correlation, while there still exists some discrepancy.
Notably, all models achieve better performance on synthetic data.
In particular, VITA has around 50\% performance improvements when switching to synthetic data.
Since CommonEval speeches are recorded with personal devices, instead of professional studio devices.
The real speeches are usually more noisy than synthetic data, which is more challenging.
These findings suggest that VITA may struggle to perform effectively during real user interactions, reinforcing our motivation to simulate real-world speech variations to thoroughly evaluate voice assistants.

\begin{table}[ht]
\centering
\resizebox{\columnwidth}{!}{ 
\begin{tabular}{ccccccc}
\toprule
\textbf{Data} & \textbf{Naïve} & \textbf{DiVA} & \textbf{LLaMA-Omni} & \textbf{Mini-Omni} & \textbf{Qwen2-Audio} & \textbf{VITA} \\
\midrule
Real & 4.04 & 3.54 &3.46   &2.02  &3.43   &2.15 \\
Synthetic &4.22  & 3.64  & 3.67  & 2.23  & 3.46   & 3.21 \\
\bottomrule
\end{tabular}
}
\caption{Performance on the real and synthetic CommonEval.}
\label{tab:synthetic-commoneval}
\end{table}

\section{Speaker Variations}
\label{sec:speaker}

\subsection{Method}
Human perception of speech is influenced by speaker-specific properties, such as accent~\citep{bradlow2008perceptual} and speaking rate~\citep{krause2004acoustic}, introducing additional complexity compared to text. 
These variations could similarly affect the performance of voice assistants. 
Motivated by this, we conduct an in-depth analysis of various speaker variations, including speaking speed, speaker age, volume, pitch, and accent, to assess their impact on voice assistants.

For speaking speed, speaker age, volume, and pitch, we perform experiments using AlpacaEval. 
Given the limited availability of instruction data from diverse speakers, we control speaking speed, volume, and pitch using Google TTS.
To obtain speeches with different speaker ages, we utilize the advanced CosyVoice-300M~\citep{du2024cosyvoice} model with speech prompts.
Our source speech prompts data includes speech recordings from the Dynamic-SUPERB age classification dataset~\citep{huang2024dynamic}, spanning speakers aged 20 to 80. 
For speakers aged 10, we collect child speech data online. 
For each age group, we select one male and one female speaker, reporting the average score across both speakers.

To examine the influence of accent, we test two settings. 
First, we use real speech samples with varying accents from the SD-QA dataset~\citep{faisal-etal-2021-sd-qa}, which includes 11 accents from regions such as Australia, the UK, North and South India, Ireland, Kenya, Nigeria, New Zealand, the Philippines, the United States, and South Africa. 
Second, we synthesize accent data using Google TTS and MeloTTS~\citep{zhao2024melo} within AlpacaEval. We report the average scores for synthetic data generated by both male and female voices using Google TTS and MeloTTS, covering accents from Australia, the UK, the United States, and India.

\begin{figure*}[!ht]
\centering
  \includegraphics[width=\textwidth]{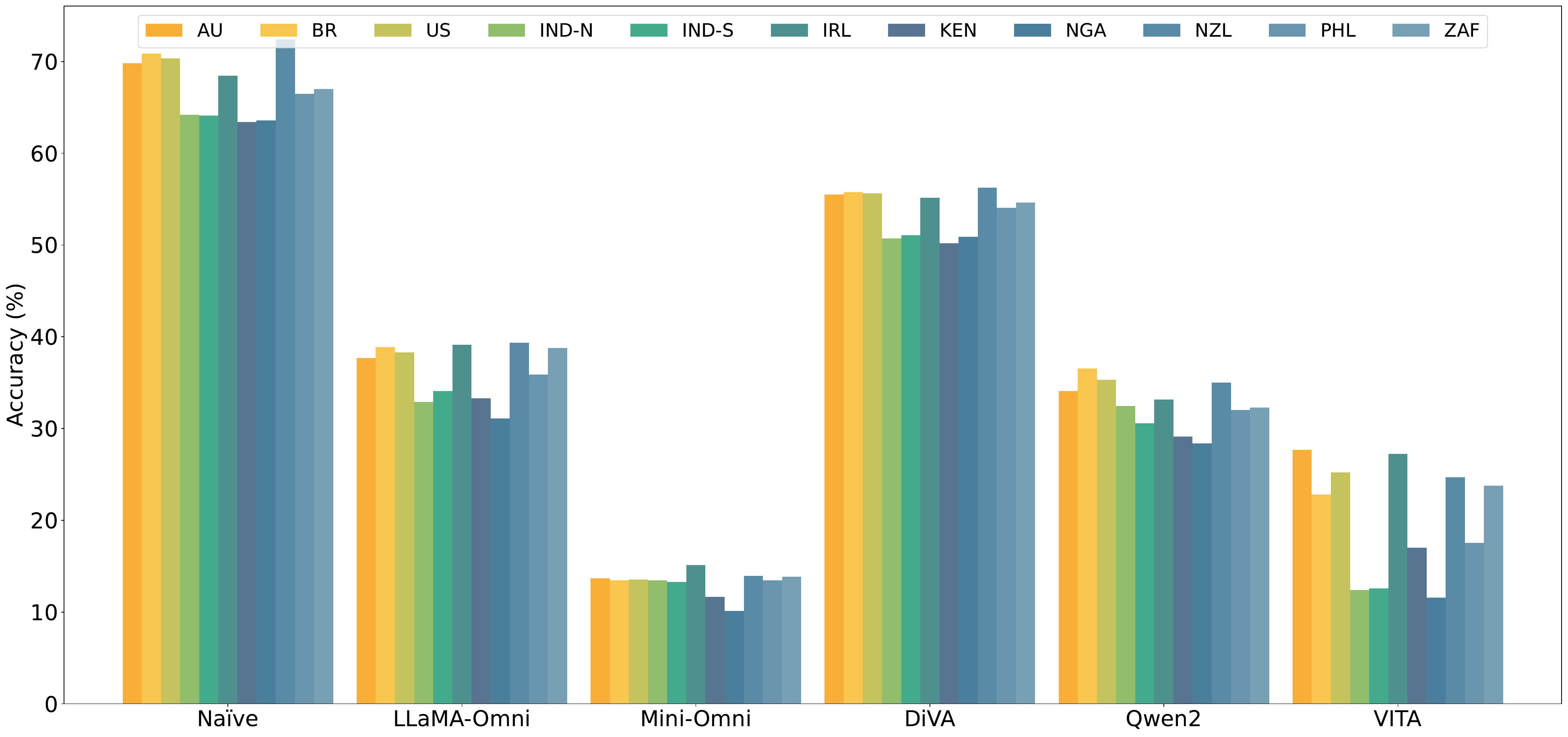} 
  \caption{Impact of accent on the performance of voice assistants (SD-QA). The locations of the accents are indicated in the figure: AU (Australia), BR (United Kingdom), US (United States), IND-N (North India), IND-S (South India), IRL (Ireland), KEN (Kenya), NGA (Nigeria), NZL (New Zealand), PHL (Philippines), ZAF (South Africa).}
  \label{fig:sd-qa}
\end{figure*}

\subsection{Results}
The impact of speaking speed, speaker age, pitch, and volume is summarized in Fig.~\ref{fig:speaker}. 
All voice assistants demonstrate a degree of resilience to speaker variations. 
The model performance is not linearly correlated with the variation level.
Instead, the model tends to maintain consistent performance in a range, and start decaying when meeting a certain threshold.
Speaker age, volume, and pitch have limited impact on model performance, with most voice assistants, except VITA, maintaining consistent performance over a wide range of variations.
Performance drops are only observed when the volume is exceptionally low. 
VITA, however, shows a degradation in performance when processing child or high-pitched speech. 
Since child speech typically has a higher pitch than adult speech, this finding is consistent with expectations.
Speaking speed has a relatively greater influence on model performance, particularly for end-to-end models. 
These models show significant degradation at speeds below 0.5x or above 1.5x, whereas the naive model remains stable across a broader range of speeds, from 0.25x to 2.0x.

\begin{figure}[ht]
\centering
  \includegraphics[width=\linewidth]{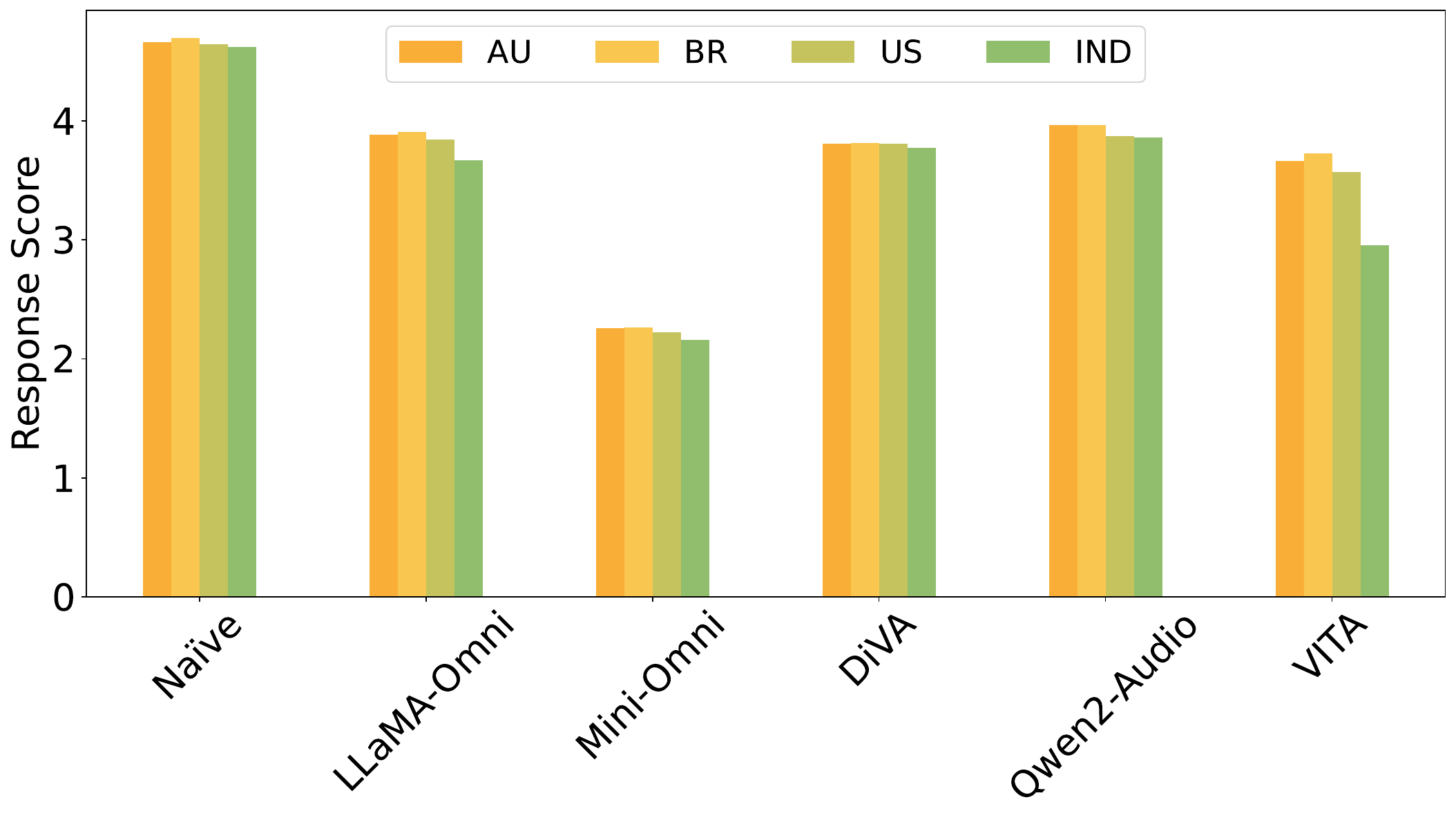} 
  \caption{Impact of accent on the performance of voice assistants (AlpacaEval).}
  \label{fig:accent}
\end{figure}

The accent results from SD-QA are presented in Fig.~\ref{fig:sd-qa}, while the accent results from AlpacaEval are summarized in Fig.~\ref{fig:accent}. 
Both synthetic and real accent results exhibit similar trends.
Overall, Mini-Omini exhibits the worst performance in response to accent variations, while Native demonstrates the best.
On high-resource accents (eg., AU, BR and US), each voice assistant shows similar performance. 
However, with low-resource accents such as Indian English and Philippines accent (IND-S, IND-N, PHL), each voice assistant experience notable performance degradation in comparison to high-resource accents. 
This degradation is more pronounced with real accent data (Fig.~\ref{fig:sd-qa}) compared to synthetic data (Fig.~\ref{fig:accent}). Given the challenges of accent TTS, synthetic accent data may not fully capture the nuanced features of accents, rendering the speech closer to standard English and thus less challenging for the models.
Similarly, VITA demonstrates more pronounced performance differences across accents compared to other voice assistants, indicating reduced robustness in handling accents, whereas the other assistants exhibit relatively consistent resilience against accent variation.
Since VITA uses a newly developed speech encoder, while the other assistants employ a Whisper-series encoder, we hypothesize that the choice of speech encoder plays a critical role in determining the robustness and generalization ability of voice assistants.

\section{Environmental Variations}
\label{sec:environment}

\subsection{Method}
When talking to voice assistants, different background environment variations pose a significant challenge to accurately understanding and responding to users' queries~\cite{Sainath2017MultichannelSP, Afouras2018TheCD, Ephraim}.
However, current evaluations lack specificity regarding noisy scenarios, which are the most common in real-world applications where voice assistants are deployed, such as in homes, vehicles, and public spaces.
To benchmark the robustness of voice assistants, we conduct a thorough analysis across various noisy conditions, including far-field speech, signal distortion, reverberation, packet loss transmission, and noise interference.

\begin{figure*}[!t]
\centering
\includegraphics[width=2\columnwidth]{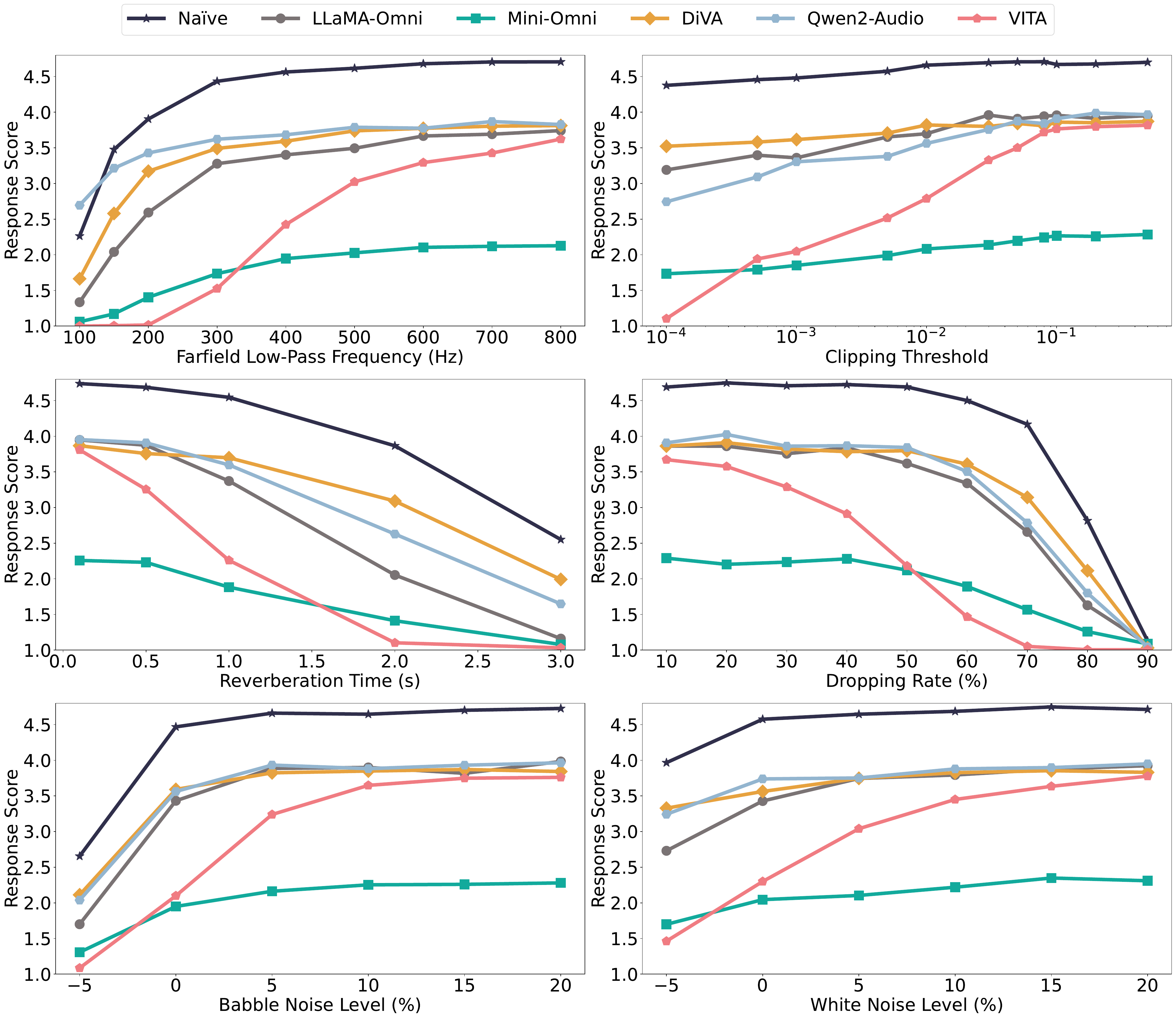} 
\caption{Impact of various environments, including far-field, speech distortion, reverberation, packet loss, babble and white noise, on the performance of voice assistants.}
\label{fig:environment}
\end{figure*}

Specifically, when the user is speaking from a distance, the speech signal weakens and high-frequency components are often attenuated due to air absorption and environmental reflection~\cite{farfield}. 
To simulate this effect, we apply low-pass filtering to the speech signal, which mimics the loss of high-frequency energy typically observed in far-field scenarios.
In real-world settings, speech distortion~\cite{distortion} can result from poor microphone quality, high-volume input, or transmission errors.
Hence, we apply clipping, which cuts off amplitude peaks that exceed a threshold, mimicking real-world over-amplification distortions and assessing the system's resilience to distorted inputs.
Reverberation occurs when speech signals reflect off surfaces, leading to overlapping echoes, particularly in large or reflective spaces. 
To simulate different levels of reverberation, we vary the reverberation time (RT60)~\cite{reverb}, which is the time it takes for sound to decay by 60 dB after the original sound stops. 
By adjusting the reverberation time, we can model environments ranging from small, acoustically treated rooms with low reverberation (short RT60) to large, echo-prone spaces like auditoriums or empty halls (long RT60).
For packet loss~\cite{packet_loss}, we simulate this by randomly dropping parts of the speech signal and assessing the assistant’s robustness in handling incomplete inputs.
Lastly, in real-world scenarios, background noise can range from steady-state noise to more complex forms of interference. 
Therefore, we simulate two common types of background noise, from steady white noise to dynamic babble noise.
Overall, by simulating different acoustic conditions, we provide a more comprehensive assessment of the voice assistant's performance in real-world applications.

\begin{table*}[!ht]
\centering
\resizebox{\textwidth}{!}{ 
\begin{tabular}{lcccccccc}
\midrule
 & \textbf{Clean} & \textbf{Repair} & \textbf{Repeat} & \textbf{Pause} & \textbf{Interjection} & \textbf{False Start} & \textbf{Misprounication} & \textbf{Grammar Error} \\
 \midrule
 Naïve & 4.68 & 4.55 (2.76) & 4.70 (-0.50) & 4.65 (0.57) & 4.59 (1.79) & 4.71 (-0.75) & 4.00 (14.47) & 4.62 (1.11) \\
DiVA & 3.86 & 3.63 (5.99) & 3.69 (4.51) & 3.63 (6.07) & 3.65 (5.34) & 3.74 (3.21) & 3.20 (17.18) & 3.75 (2.86) \\
LLaMA-Omni & 3.95 & 3.24 (17.94) & 3.84 (2.80) & 3.89 (1.61) & 3.88 (1.78) & 3.86 (2.16) & 3.22 (18.58) & 3.82 (3.27) \\
Mini-Omni & 2.25 & 1.85 (18.13) & 1.95 (13.67) & 1.84 (18.42) & 1.97 (12.41) & 2.20 (2.30) & 1.83 (19.02) & 2.19 (2.90) \\
Qwen2-Audio & 3.89 & 3.49 (10.41) & 3.34 (14.19) & 3.36 (13.85) & 3.36 (13.85) & 3.77 (3.18) & 3.11 (20.22) & 3.83 (1.76) \\
VITA & 3.78 & 2.85 (24.61) & 2.82 (25.45) & 2.83 (25.14) & 2.73 (27.89) & 3.40 (10.05) & 2.51 (33.60) & 3.61 (4.65) \\
\midrule
Avg. & 3.74 & 3.27 (12.55) & 3.39 (9.30) & 3.36 (9.95) & 3.36 (9.97) & 3.62 (3.26) & 2.98 (20.34) & 3.64 (2.68) \\
\bottomrule
\end{tabular}
}
\caption{The impact of content noise on the performance of voice assistants. For each cell, we show the response score and performance degradation percentage after injecting content noise.}
\label{tab:content-results}
\end{table*}

\subsection{Results}
The performance of voice assistants under different environmental conditions are summarized in Fig.~\ref{fig:environment}.
Similar to the observations of speaker variations, all models equipped with whisper speech encoders demonstrate similar levels of resilience across various conditions.
Notably, the naive pipeline model outperforms the others, while the VITA model shows significantly lower performance.
A plausible explanation for this trend is that the cascaded model, i.e., naive model, first transcribes the speech into text before passing it to LLM, potentially providing an implicit denoising process.
This intermediate transcription step may help mitigate noise and enhance the clarity of input, benefiting the following LLM's performance.
While for the end-to-end models, i.e., LLaMA-Omni, Mini-Omni, DiVA, Qwen2-Audio, and VITA, process the audio signal directly, as a result, they are more susceptible to noisy conditions.

\section{Content Variations}
\label{sec:content}

\subsection{Method}
Compared to written text, spoken language tends to be more informal and casual, often containing various errors such as disfluencies~\citep{TREE1995709,shriberg1994preliminaries,jamshid-lou-johnson-2020-end,marie-2023-disfluency}, mis-pronunciation~\citep{kheir-etal-2023-automatic,dell1981stages}, and grammar error~\citep{carter1995grammar,mccarthy1995spoken,caines-etal-2020-grammatical}.
These errors are common in natural speech and can have a significant impact on the performance of voice assistants. However, current evaluations of voice assistants often focus on clean data, overlooking these frequent speech errors.
In this section, we analyze the effects of common speech content errors on voice assistant performance. Specifically, we examine mispronunciations, grammatical errors, and a range of common disfluencies, including repairs, repetitions, filled pauses, interjections, and false starts.
Typical examples of each content error type are provided in Tab.~\ref{tab:content-example}. Due to the scarcity of instruction data containing such errors, we leverage GPT-4o to rewrite clean instructions into noisy versions using a few-shot demonstration approach. The instructions from AlpacaEval are rewritten to include various types of errors, and the modified text is then converted into speech using Google TTS.

\begin{table}[ht]
\centering
\small
\begin{tabularx}{\columnwidth}{lX}
\toprule
Clean & What's the placebo phenomenon? \\
\midrule
Repair & What's \textbf{the nocebo... I mean,} the placebo phenomenon? \\
Repeat & What's the \textbf{the the} placebo phenomenon? \\
Pause & What's, \textbf{uh}, the placebo phenomenon? \\
Interjection & \textbf{Well}, what's the placebo phenomenon? \\
False start & \textbf{I was thinking...} What's the placebo phenomenon? \\
Misprounication & What's the placebo phenome\textbf{m}on? \\
Grammar error & What's the placebo phenomenon \textbf{is}? \\
\bottomrule
\end{tabularx}
\caption{Examples of different content variations.}
\label{tab:content-example}
\end{table}

\subsection{Results}
The performance of voice assistants under various content errors is summarized in Tab.~\ref{tab:content-results}.
Overall, naive pipeline models demonstrate the best robustness in handling content errors. 
All voice assistants show strong resilience to grammatical errors but are much more vulnerable to mispronunciations.
Mispronunciations often result in a large number of incorrectly recognized words, leading to a higher Word Error Rate (WER) in the transcription, which can alter the intended meaning of the speech. 
In contrast, grammatical errors tend to preserve the overall meaning, leading to less disruption in performance.
Additionally, LLMs also display a high tolerance for grammatical errors but exhibit much less resilience to high WER, likely because grammatical mistakes are common in written text, while incorrect word recognition is not~\citep{wang2024resilience}. 
The vulnerability of base LLMs can also help explain the observed trends in performance.
Similarly, spoken disfluencies—commonly absent in written text—significantly degrade model performance. Disfluencies can be considered as including irrelevant content, which can easily distract the LLMs~\citep{pmlr-v202-shi23a}.
Among the different types of disfluencies, repairs are the most problematic. 
Repairs can introduce incorrect or conflicting information that causes the model to misinterpret the query and produce incorrect responses. 
For instance, as shown in Tab.~\ref{tab:content-example}, the voice assistant might incorrectly respond to the definition of "nocebo" instead of "placebo." 
Consequently, repair disfluencies cause the greatest performance degradation compared to other types of disfluencies.

\section{Conclusion}
In this work, we introduce the first comprehensive multi-facet benchmark to assess the capabilities of voice assistants using both real and synthetic spoken instructions. Our results highlight a significant performance gap between end-to-end models and straightforward pipeline models, underscoring the need for further advancements in processing spoken instructions effectively.
Additionally, we uncover key vulnerabilities in voice assistants by evaluating their performance across various factors, including speaker variations, environmental conditions, and content-related errors. These findings suggest important areas for improvement in voice assistant robustness.
Future work includes developing evaluation protocols for speech-based responses and extending the benchmark to incorporate more diverse and realistic evaluation data..


\bibliography{anthology,custom}
\clearpage
\appendix

\section{Generation Prompts}
\label{appx:generation-prompt}
The data generation prompts used in our experiments are listed in this section.
\begin{itemize}
    \item \textbf{Text normalization:}
\begin{tcolorbox}
\footnotesize
\texttt{You are tasked with normalizing text for a Text-to-Speech (TTS) system. Your job is to take a raw text input and transform it into a form that a TTS engine can easily process. This includes:}

\texttt{1. Expanding abbreviations, acronyms, and contractions.}

\texttt{2. Converting numbers into their word forms.}

\texttt{3. Expanding dates, times, and units of measurement into their spoken equivalents.}

\texttt{4. Handling special characters (such as "\$", "\#", "*", ">", "<", "\textbackslash n", "-") and ensuring they are converted into their words equivalents.}

\texttt{5. Correctly formatting currency, percentage, and other symbols.}
\texttt{6. Preserving proper names and specific phrases but normalizing other text elements.}
\newline

\texttt{Here’s the text to normalize:}

\texttt{Text: [[instruction]]}
\newline

\texttt{Please output the normalized instruction only without anything else!}
\end{tcolorbox}

\end{itemize}

\section{Evaluation Prompts}
\label{appx:eval-prompts}

The automatic evaluation prompts used in our experiments are listed in this section.

\begin{itemize}
\item \textbf{SD-QA:}
\begin{tcolorbox}
\footnotesize
\texttt{\#\#\# Question}

\texttt{[[Question]]}
\newline

\texttt{\#\#\# Reference answer}

\texttt{[[Answer]]}
\newline

\texttt{\#\#\# Candidate answer}

\texttt{[[Response]]}
\newline

\texttt{Is the candidate answer correct based on the question and reference answer?}

\texttt{Please only output a single "Yes" or "No". Do not output anything else.}
\end{tcolorbox}

\item \textbf{AlpacaEval \& CommonEval:}

\begin{tcolorbox}
\footnotesize
\texttt{I need your help to evaluate the performance of several models in the speech interaction scenario. The models will receive a speech input from the user, which they need to understand and respond to with a speech output.}

\texttt{Your task is to rate the model’s responses based on the provided user input transcription [Instruction] and the model’s output transcription [Response].}
\newline

\texttt{Please evaluate the response on a scale of 1 to 5:}

\texttt{1 point: The response is largely irrelevant, incorrect, or fails to address the user’s query. It may be off-topic or provide incorrect information.}

\texttt{2 points: The response is somewhat relevant but lacks accuracy or completeness. It may only partially answer the user’s question or include extraneous information.}

\texttt{3 points: The response is relevant and mostly accurate, but it may lack conciseness or include unnecessary details that don’t contribute to the main point.}

\texttt{4 points: The response is relevant, accurate, and concise, providing a clear answer to the user’s question without unnecessary elaboration.}

\texttt{5 points: The response is exceptionally relevant, accurate, and to the point. It directly addresses the user’s query in a highly effective and efficient manner, providing exactly the information needed.}
\newline

\texttt{Below are the transcription of user’s instruction and models’ response:}

\texttt{\#\#\# [Instruction]: [[Instruction]]}

\texttt{\#\#\# [Response]: [[Response]]}

\texttt{After evaluating, please output the score only without anything else.}

\texttt{You don’t need to provide any explanations.}
\end{tcolorbox}

\end{itemize}

\section{Evaluation results with different TTS models}
\label{appx:tts-results}

The evaluation results on synthetic speeches generated by various TTS models are summarized in Tab.~\ref{tab:tts-results}. Overall, the voice assistants demonstrate the best performance on data produced by Google TTS, highlighting the superior quality of Google’s text-to-speech system. This outcome reinforces the effectiveness of Google TTS in generating realistic and high-quality synthetic speech. Furthermore, we observe a consistent ranking of models across different synthetic datasets, which underscores the reliability and validity of using Google TTS synthetic data as a benchmark for evaluating the performance of voice assistants.

\begin{table}[ht]
\centering
\resizebox{\columnwidth}{!}{ 
\begin{tabular}{lcccccc}
\toprule
& \textbf{Text} & \textbf{CozyVoice-M} & \textbf{CozyVoice-F} & \textbf{Google-M} & \textbf{Google-F} & \textbf{MeloTTS} \\
\midrule
Naive   & 4.81  & 4.43 & 4.56 & 4.68  & 4.73  & 4.52 \\
DiVA& 4.84  & 3.60 & 3.71 & 3.86  & 3.85  & 3.81 \\
LLaMA-Omni  & 4.58  & 3.68 & 3.73 & 3.95  & 4.03  & 3.54 \\
Mini-Omni   & 2.64  & 2.17 & 2.22 & 2.25  & 2.30  & 2.11 \\
Qwen2-Audio & 4.27  & 3.58 & 3.73 & 3.89  & 3.96  & 3.76 \\
VITA& 4.16  & 3.21 & 3.49 & 3.78  & 3.79  & 3.14 \\
\midrule
Avg.& 4.22  & 3.45 & 3.57 & 3.74  & 3.78  & 3.48 \\
\bottomrule
\end{tabular}
}
\caption{Performance of voice assistants on AlpacaEval generated with different TTS systems. M refers to male voice, and F refers to female voice.}
\label{tab:tts-results}
\end{table}

\end{document}